\documentclass{article}
\usepackage[utf8]{inputenc}
\usepackage{comment}
\usepackage{hyperref}
\usepackage{graphicx}
\usepackage{amsmath,amssymb}
\usepackage{xcolor}
\definecolor{light-gray}{gray}{0.95}
\newcommand{\code}[1]{\colorbox{light-gray}{\texttt{#1}}}

\title{\textbf{M-VADER: A Model for Diffusion with Multimodal Context}}

\author{
\normalsize{}
\textbf{Samuel Weinbach$^1$ \thanks{Correspondence: \texttt{{\{samuel.weinbach, marco.bellagente\}@aleph-alpha.com}}}} \hspace{8mm} 
\textbf{Marco Bellagente$^1$ }\footnotemark[1] \\
\normalsize{}
\textbf{Constantin Eichenberg$^1$} \hspace{8mm}
\textbf{Andrew Dai$^1$} \\
\normalsize{}
\textbf{Robert Baldock$^1$} \hspace{8mm}
\textbf{Souradeep Nanda$^1$} \\
\normalsize{}
\textbf{Bj{\"o}rn Deiseroth$^{1,2}$} \hspace{8mm}
\textbf{Koen Oostermeijer$^1$} \\
\normalsize{}
\textbf{Hannah Teufel$^1$} \hspace{8mm}
\textbf{Andres Felipe Cruz-Salinas$^1$} \\
\normalsize{}
$^1$Aleph Alpha GmbH, Heidelberg, Germany\\
\normalsize{}
$^2$Artificial Intelligence and Machine Learning Lab, TU Darmstadt, Germany
}

\date{}

\begin{document}

\maketitle

\begin{abstract}
We introduce M-VADER\footnote{The name was suggested by the model itself}: a diffusion model (DM) for image generation where the output can be specified using arbitrary combinations of images and text.
We show how M-VADER enables the generation of images specified using combinations of image and text, and combinations of multiple images.
Previously, a number of successful DM image generation algorithms have been introduced that make it possible to specify the output image using a text prompt.
Inspired by the success of those models, and led by the notion that language was already developed to describe the elements of visual contexts that humans find most important, we introduce an embedding model closely related to a vision-language model.
Specifically, we introduce the embedding model S-MAGMA: a 13 billion parameter multimodal decoder combining components from an autoregressive vision-language model MAGMA and biases finetuned for semantic search.
\end{abstract}

\section{Introduction}

\begin{figure}[h!]
\noindent\makebox[\textwidth]{
    \centering
    \includegraphics[scale=0.45]{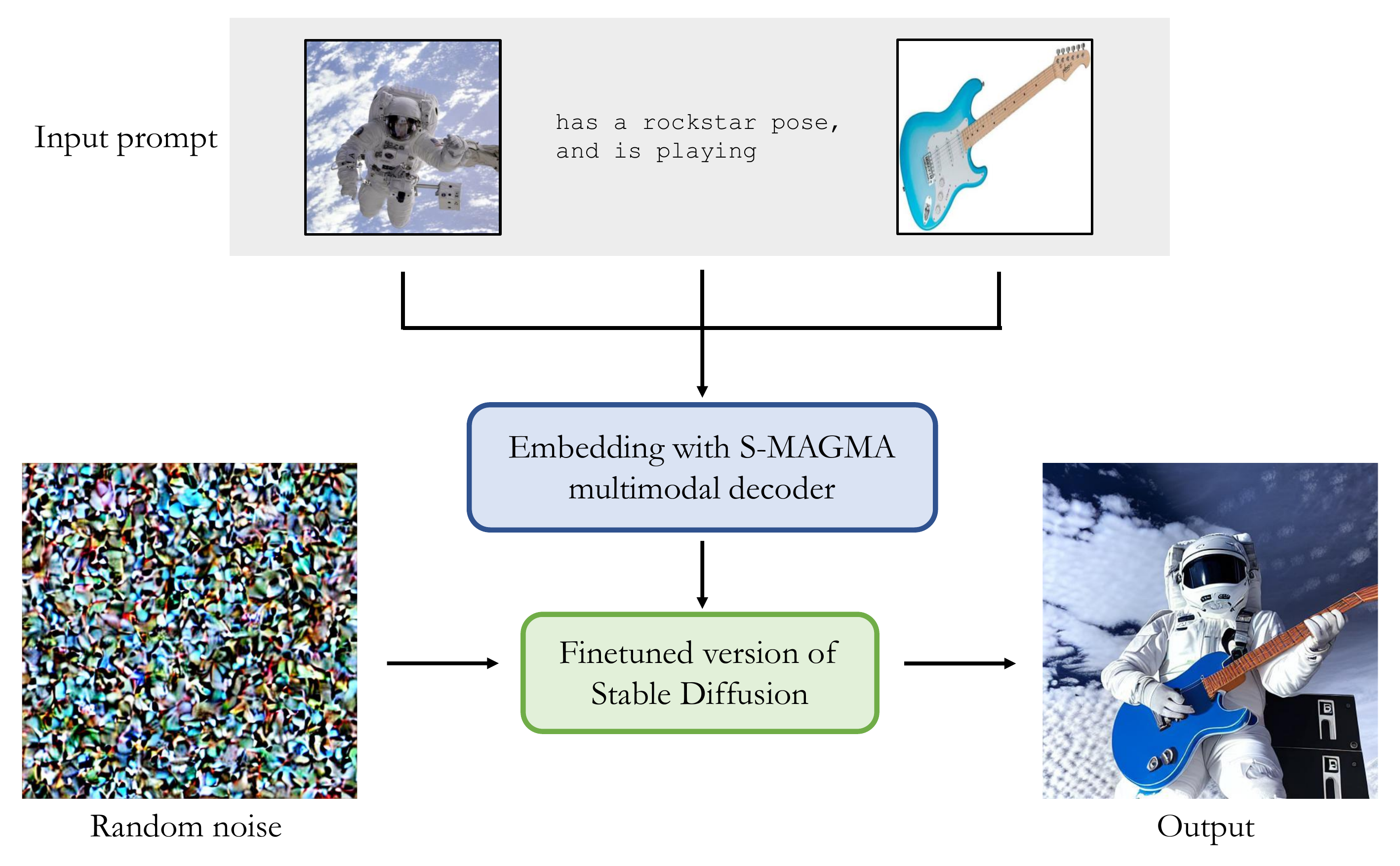}
}
\caption{M-VADER: image synthesis with multimodal context. The guidance prompt, comprised of interleaved images and text, is embedded using a multimodal decoder, S-MAGMA. The output of S-MAGMA is used to condition the generation process of a fine-tuned version of Stable Diffusion via cross-attention. This allows the finetuned version of Stable Diffusion to convert the starting (random noise) image into the output image shown.}\label{fig:figure_1}
\end{figure}
In recent years, advances in text-to-image synthesis have enabled the creation of artistic and photorealistic images with impressive generalization from text prompts \cite{high_res_diffusion, glide, dall_e_2, imagen, parti}.
In particular, text guided diffusion models have been developed, which transform either random noise, or a starting image, into a new image, guided by a text contextual prompt.
The natural question to ask is how one can guide diffusion models using multimodal guidance contexts.

Recently, a version of Stable Diffusion~\cite{high_res_diffusion} has been developed that uses single images as conditional information for the diffusion process \cite{single_image_guided_stable_diffusion}. However, there is no effective method to use multimodal context guidance in diffusion models: a context that combines one or more passages of text with one or more images. The richness in information contained in the desired image output creation is sometimes hard to capture with a (single modal) text prompt alone.
This is the key contribution of this paper: a method for image generation with multimodal contextual guidance (of arbitrary prompt length). 
To this end, we develop a novel embedding model for multimodal contextual prompts, and we finetune Stable Diffusion to follow contextual guidance.

M-VADER is composed by assembling components from two 13B trans\-former-based models: a MAGMA
image captioning model~\cite{magma}, and a 13B parameter ``Luminous-Explore'' symmetric semantic search model~\cite{explore, sgpt}. We additionally show how to control the relative importance of the components of the multimodal guidance prompt, by reweighting the attention scores to each component. 
Our contributions are therefore:
\begin{itemize}
    \item S-MAGMA, a multimodal decoder embedding model
    \item An image generation diffusion model, which we call M-VADER, (finetuned from Stable Diffusion), which follows multimodal guidance
    \item A method for scaling the relative importance of individual components in the multimodal guidance prompt, leading to qualitatively predictable changes in the images generated by M-VADER
\end{itemize}

Using M-VADER we demonstrate the following capabilities:
\begin{itemize}
    \item Image generation guided using multimodal context
    \item Image composition, generation guided using the combination of two images
    \item Image variation, by transforming random noise into a final image, guided using an image to be varied.
\end{itemize}
\section{Related Work}
Various approaches exists to generative modelling of high dimensional distributions.
Generative adversarial networks (GANs) \cite{gan, vqgan_clip} are likelihood-free models whose objective is to minimize a sample-based distance with respect to the target distribution. Image generation with GANs scales efficiently with the pixel space size, while also being able to capture the bulk of distributions. However, they struggle at modelling distribution tails and their training dynamic is often unstable.

Conversely, likelihood-based models minimize the negative log-likelihood with respect to the target distribution either in pixel space or a compressed space. They can be realized as amortized inference models~\cite{vae, vq_vae2}, bijections~\cite{flow1, flow2} or autoregressive models (ARMs) based on chain-rule decomposition into conditional probabilities~\cite{autoregressive1, autoregressive2, autoregressive3}. Within the family of likelihood-based models, ARMs achieve the best results in terms of density estimation. However, modelling images directly in pixel space with ARMs scales poorly with the pixel size, limiting such models to low resolution images. This scaling issue has been solved by getting rid of the pixel space representation of images and rather compress them to a latent space which is then modelled by an ARM~\cite{taming, vq_vae2, vqgan}.

Diffusion models have seen a renaissance in recent years. Originally introduced in \cite{ddpm3}, they went largely ignored in favour of other generative methods such as variational autoencoders (VAEs) \cite{vae} and GANs. In 2020, interest picked up again after Ho et al. published a state-of-the-art diffusion model and a simplified training objective \cite{ddpm}. It was found that diffusion models outperform GANs in terms of sample quality in unconditional generation \cite{ddpm2}.

A diffusion model is a generative model defined by a forward and backward process. In the forward process noise is gradually added to the data until it becomes indistinguishable from the prior distribution, typically an isotropic Gaussian distribution. The model is trained to reverse this noise corruption in the backward process, after which it can be used to generate new samples. It can be thought of as a hierarchical VAE, consisting of a chain of latents, each corresponding to an increasing level of noise. Like a VAE, it is optimised with an evidence lower-bound (ELBO) loss objective, however, the encoder is pre-defined. This prevents mode collapse, caused by the simultaneous training of two networks, which have historically plagued GANs and standard VAEs.

They are also shown to be equivalent to denoising score-based methods \cite{ddpm}, which use the gradient of the logarithm of the data distribution and Langevin dynamics for estimation \cite{du2019implicit, score2, song2019generative}. The continuous-time version is defined as a stochastic differential equation (SDE), it can be shown that all the different diffusion models are discretizations of this more general framework \cite{karras2022elucidating}.

A drawback of diffusion models is that traversing the whole Markov chain in the backward process would require thousands of model calls, resulting in sampling speeds that are orders of magnitude slower than other methods. This shortcoming was alleviated by a strided method in which multiple steps are taken at once \cite{nichol2021improved}. From the SDE perspective following the backward process is equivalent to solving the SDE. This insight has allowed for other sampling methods to be introduced, based on SDE and ODE solvers, leading to further improvements \cite{karras2022elucidating, lu2022dpm}.

Dhariwal and Nichol \cite{nichol2021improved} introduced a method whereby the diffusion process could be steered through the use of a classifier as a way to realize conditional generation. To avoid having to rely on a classifier, it is also possible to train the model on a dataset where part of the images have text labels. The same model can then be used to do both conditional and unconditional generation \cite{classifierfree}. This has become a standard tool for text-to-image diffusion models \cite{glide, dall_e_2, imagen, high_res_diffusion}. Our work extends the conditional generation method used in these prior works to multimodal conditioning, allowing for a new, more-flexible way to steer image generation.

Besides diffusion models, autoregressive generative models such as Parti \cite{parti} have yielded comparable results.

\begin{figure}[h!]
    \includegraphics[scale=0.8]{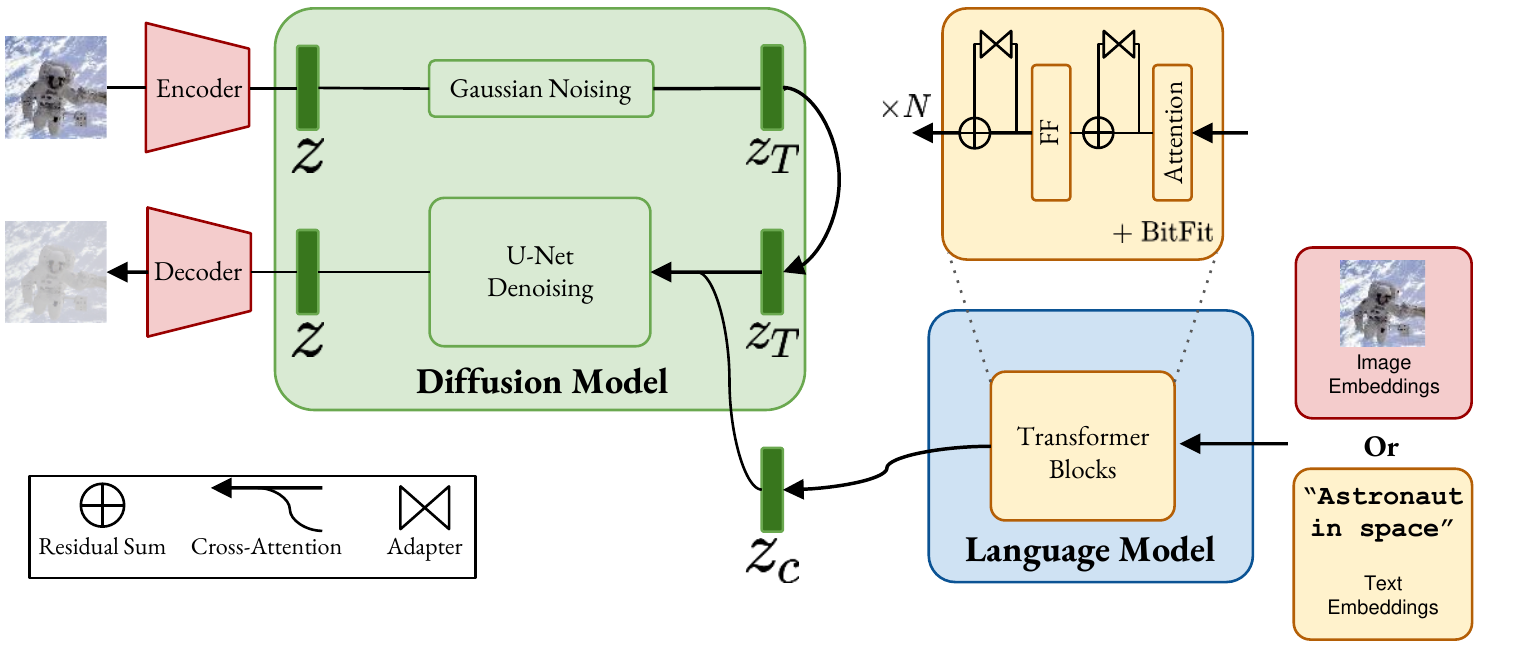}
\caption{M-VADER architecture for training. We condition the denoising model through cross attention with embeddings from a Decoder Language Model augmented with multimodal components (MAGMA) and finetuned biases for semantic search. In training we use either the image or the caption for conditioning, while inference is performed with multimodal input.
\textit{Notation} - $z$: Image latent. $z_T$: Noisy latent. $z_c$: Conditioning latent.}
\label{fig:architecture}
\end{figure} 
\section{Method}
In this section we describe the model architecture and discuss the details of training. Then we present our new \textit{multimodal prompting} paradigm for inference which makes use of attention manipulation. 
\subsection{Architecture} \label{subsec:architecture}

The general architecture consists of three main components: A pre-trained LLM; finetuned prefix and bias components enabling multimodal input and semantic embeddings; and a diffusion model to generate image output which is conditioned through embeddings produced by the LM stack.  We describe these components in detail. \\

\textbf{Language model.}
Embeddings from a pre-trained LM can be employed effectively to condition diffusion models, and scaling the LM backbone is an effective way to achieve better quality and alignment in generation~\cite{imagen}. For this work we use the 13B parameter model from Aleph Alpha's Luminous family trained on $\sim$400B language tokens. Luminous largely follows the architecture of GPT-3, the major difference being the use of rotary positional embeddings in our model. Luminous 13B is trained on a curated multilingual corpus containing sources in English, German, French, Italian and Spanish. \\

\textbf{Finetuned components.}
It has been shown in \cite{magma, frozen, flamingo} that the few-shot capabilities of LMs can be extended to multimodal prompts consisting of arbitrary combinations of images and text. Following the architecture introduced in MAGMA~\cite{magma}, we extend Luminous 13B to a multimodal model by adding a CLIP Resnet encoder~\cite{clip}, a prefix layer and adapters which are then finetuned on an autoregressive captioning task. Independent of multimodal pre-training we also finetune the biases (c.f. BitFit~\cite{bitfit}) of Luminous 13B on a semantic contrastive objective~\cite{sgpt}. These biases can be combined in plug-and-play fashion with the multimodal components and yield surprisingly good results on image-text similarity tasks, implying that the semantic feature extraction from the biases extends to multimodal MAGMA embeddings. We stress that both finetuning procedures leave the weights of the LM completely frozen, thereby keeping the world knowledge of the model intact and preventing catastrophic forgetting. In the following we refer to the stack of Luminous 13B together with its finetuned multimodal and semantic components as Luminous S-MAGMA. \\

\textbf{Diffusion model.}
The final component is the U-Net-based Denoising Diffusion Probabilistic Model (DDPM) from Stable Diffusion\footnote{https://github.com/CompVis/stable-diffusion}\footnote{While finalizing this work, a new version of Stable Diffusion has been released https://stability.ai/blog/stable-diffusion-v2-release. The starting checkpoint as well as all results refer only to version 1.4}.
The diffusion model takes an image input and a conditional input, which can both be empty, and outputs an image\footnote{in Stable Diffusion, the diffusion process happens in a compressed latent space of a pre-trained VAE, hence image input/output needs to be additionally encoded/decoded. For readability, we mostly suppress this in our notation.} by applying a denoising sampling procedure resembling Langevin dynamics. Given an image and conditional input (in the case of Stable Diffusion, a natural language prompt embedded by CLIP's text encoder), gradual noise is added to the image to create a latent vector. Then a new image is generated by applying the denoising process to the noised latent while using the conditional input to steer the generation through classifier-free guidance~\cite{classifierfree}. If no image input is given, the latent vector is sampled from Gaussian noise, and if no condition is given the denoising process is applied without any bias. \\

\textbf{End-to-end architecture.} The formulation of DDPMs allows the generation process to be conditioned on a variety of modalities of inputs such as text prompts or bounding boxes. In Fig.~\ref{fig:architecture} we provide an illustration of the model. In our architecture, we leverage our augmented decoder model as follows. Given any multimodal input, i.e. a sequence of interleaved images and texts, we perform a forward pass through Luminous S-MAGMA to obtain a sequence of hidden states after the final transformer layer. These hidden states are then used to condition the unet through a cross-attention layer in each of the unet's attention layers~\cite{high_res_diffusion, imagen}. Note that the different finetuned components of S-MAGMA can be plugged in and out in this approach, and be finetuned or frozen during training. This allows for interesting combinations that have a confounding effect on the quality, but also the style of the generations, see Section~\ref{sec:results}.

\subsection{Training} \label{subsec:training}

\textbf{Training objective.} We follow standard practice and train the diffusion model, denoted by $F_\theta(x, c, t)$, to predict the noise $\epsilon \sim \mathcal{N}(0,1)$ from a corrupted sample $\tilde{x}(\epsilon, t)$
via MSE loss at a randomly sampled diffusion time step $t \sim \mathcal{U}(0, T)$,

\begin{align}
\begin{split}
    &\mathcal{L}_\theta = \mathbb{E}_{t, \epsilon, x} || \epsilon - F_\theta(\tilde{x}(\epsilon, t), c, t) ||^2. \\ 
\end{split}
\end{align}

Here, $x$ is the image input and $c$ is the conditional input. We train on image caption pairs $(x_{img}, x_{txt})$ and denote by $H(x)$ the sequence of hidden states after the forward pass through Luminous S-MAGMA, detailed in the previous subsection. For each image caption pair during training we use either $c = H(x_{img})$ or $c = H(x_{txt})$ for conditioning the diffusion model, with probability of using the image being $0.2$. More training details can be found in Tab.~\ref{tab:training_diffuser}. \\
\begin{figure}[t!]
    \centering
    \includegraphics[scale=0.6]{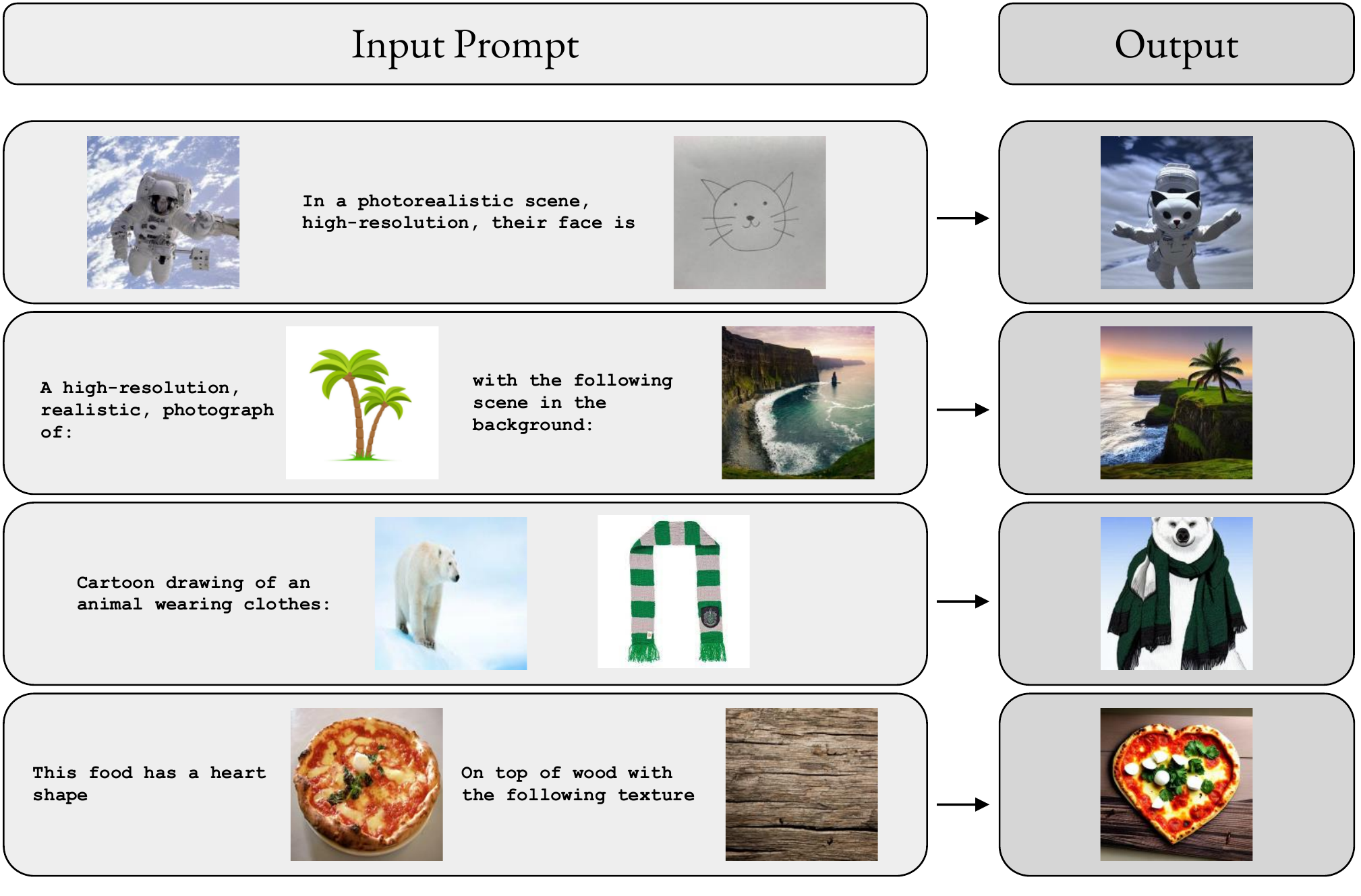}
\caption{Examples of image generation with multimodal guidance. Output images (right) are composed from the combination of interleaved images and text shown on the left.}\label{fig:multimodal_prompt}
\end{figure} 
\textbf{Trained components}. State-of-the-art DDPM models are pre-trained for millions of steps on large batch sizes~\cite{high_res_diffusion, dall_e_2, imagen}. Given the availability of the public checkpoint from Stable Diffusion, an efficient strategy is to only train from scratch the newly added parameters. In this work, these are the cross-attention parameters needed to accommodate the multimodal hidden states. There is also the option to unfreeze and finetune various components from S-MAGMA, meaning the embedding preprocessor $H = H_\theta$ would be updated during training as well. There are numerous possible ablations using (or not using) the components of MAGMA or Semantic Search, from which we have trained 2 models so far, see Tab.~\ref{tab:training_variants}. \\

\textbf{Dataset}. We train the model on LAION aesthetics V.2 5+\footnote{https://laion.ai/blog/laion-aesthetics/}, i.e. the subset of LAION 5B filtered by aesthetic score with a predicted aesthetic score $> 5$~\cite{laion}. Filtering only images with a minimum resolution of $512\times512$, this results in roughly $40$M captioned images.\\

\subsection{Multimodal prompting}
In inference we synthesize images using a variety of conditioning prompts consisting of interleaved visual and textual data
\begin{align*}
    \code{prompt: [<Text$_1$>, <Image$_1$>, <Text$_2$>, <Image$_2$>, \ldots].}
\end{align*}
Any prompt $x$ of the above format is converted into a conditional embedding $c = H(x)$ and we utilize classifier free guidance to generate an image from a pure Gaussian noise vector. Throughout our experiments we use a guidance scale of 8.0 and the denoising differential equation is solved in 50 steps with the pseudo-numerical method described in~\cite{pndms}. Because of the flexible format, multimodal prompting can be used for quite complex image synthesis tasks, see Fig.~\ref{fig:multimodal_prompt} for some examples. Two specific tasks where this method excels are style change of images and image composition, which we elaborate on in Section~\ref{sec:results}. \\
An important ingredient for successful multimodal prompting is attention manipulation, which we explain below.

\subsection{Attention manipulation}

By MAGMA's architecture, every image translates into a sequence of $144$ embeddings, whereas the caption typically is between $10$ and $20$ tokens long. Because of this, for conditional prompts of the form $[\mathrm{image}, \mathrm{text}]$ (or the other way around) the image is implicitly upweighted in the cross-attention layers, making multimodal prompting difficult out of the box. To counteract this phenomenon, during inference we modify the attention scores in every attention layer of the transformer as follows. \\
Let $S = (s_{ij})$ denote the pre-softmax attention scores inside an attention head. Let $\theta_i$ be a non-negative number for every token $i$. Then we obtain the modified scores $\tilde{S} = (\tilde{s}_{ij})$ by the simple operation

\begin{align} \label{e:atman}
    \tilde{s}_{ij} = s_{ij} + \log \theta_i,
\end{align}
meaning that $\theta_i > 1$ upweights the $i$-th token in the self-attention, whereas $\theta_i < 1$ downweights it, with $\theta_i = 0$ completely suppressing token $i$ after the softmax operation. \\
For a prompt $[\mathrm{image}, \mathrm{text}]$ we found that it is usually beneficial to downweight all image tokens uniformly, meaning we would set $\theta_i = \rho < 1$ for $i \leq 144$ and $\theta_i = 1$ else. Attention manipulation also helps with prompts containing two images $[\mathrm{image}_1, \mathrm{image}_2]$, where we found that sometimes one image has a much more dominant effect on the diffusion process than the other one, and attention manipulation can be used to balance the two effectively. \\
In theory, different attention factors across layers and/or attention heads could be used, but we kept it simple with $\theta_i$ being the same in every attention operation.
\begin{figure}[t]
    \centering
    \includegraphics[scale=0.35]{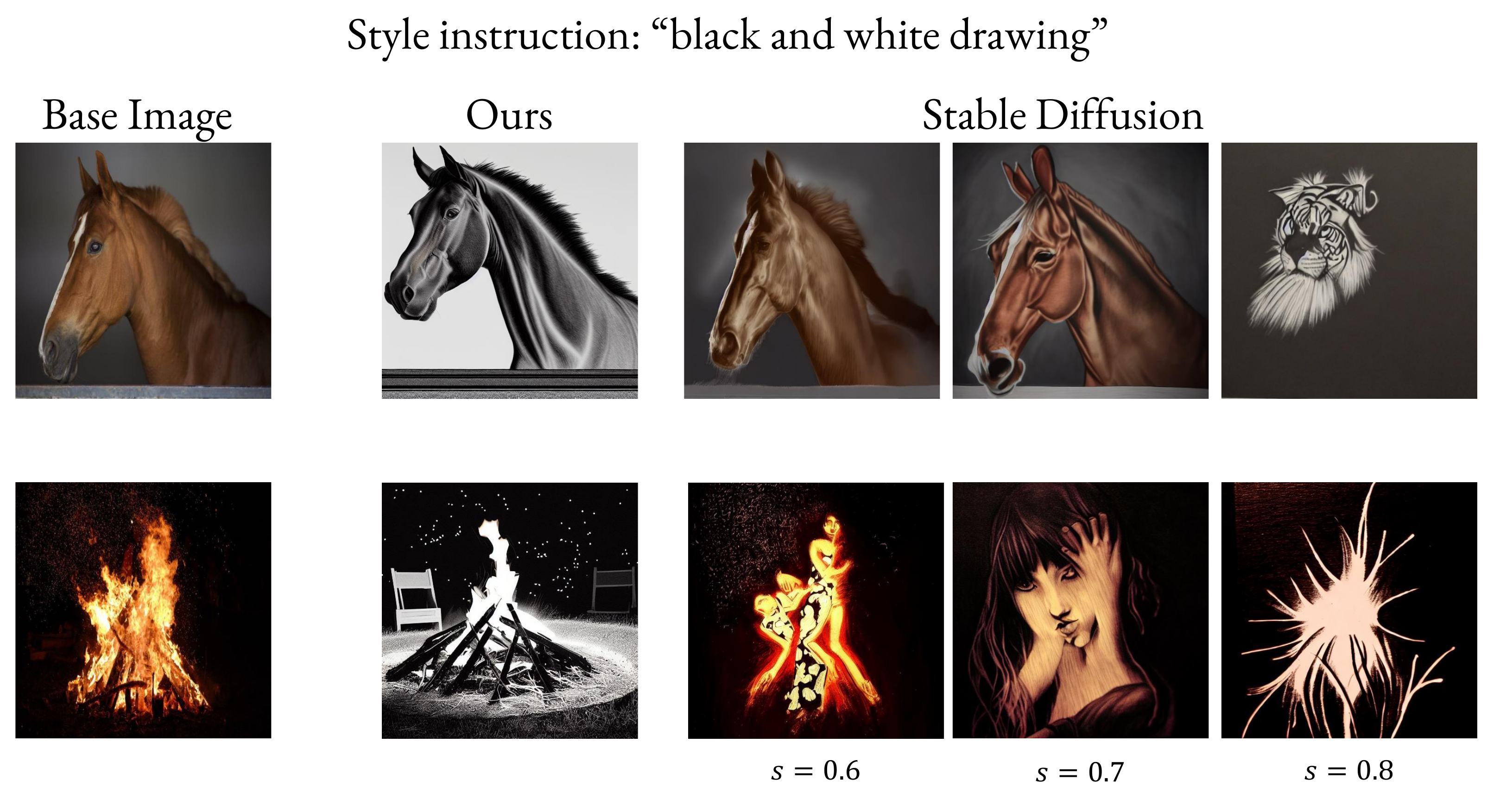}

\caption{Style modification of a base image. For Stable Diffusion, we begin the diffusion process from the noised base image and guide the diffusion process with a text prompt. The \textit{strength} parameter $s$ denotes the relative level of input noise, where $s=0$ and $s=1$ correspond respectively to no noise and full noise levels. We compare the outputs of this approach with varying strengths to the M-VADER multimodal prompt which always generates from pure noise.}\label{fig:multimodal-style}
\end{figure}
\section{Results} \label{sec:results}
In this section we highlight the capabilities of M-VADER and multimodal prom\-pting. A number of examples are given in Fig.~\ref{fig:multimodal_prompt}, demonstrating meaningful composition of several concepts (captured in our multimodal embeddings) into a generated output image that aligns with the input condition more than a purely textual condition. We mainly focus on the image synthesis tasks of style modification and image composition. As it is not the goal of this work to improve the original model of~\cite{high_res_diffusion} on pure text conditioning, we only include such examples in Appendix~\ref{sec:watermarks}. We also remark that our model offers an alternate approach to the task of image variation by prompting with just a single image and include examples of this in Appendix~\ref{app:B}. We remark that these are purely qualitative results, as a proper quantitative evaluation is currently not feasible (see Section~\ref{sec:limitations}).

\subsection{Style modification}
\begin{align*}
    \code{prompt: [<Text$_1$>, <Image$_1$>, \ldots].}
\end{align*}

\begin{figure}[t]
    \centering
    \includegraphics[scale=0.65]{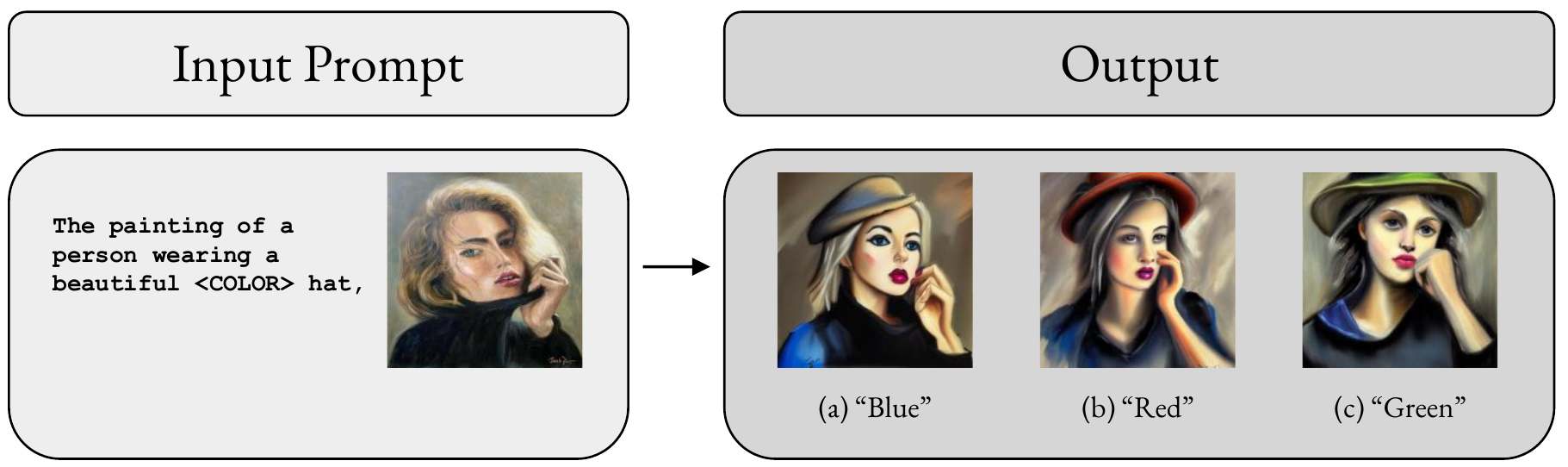}
\caption{M-VADER allows image variation with the addition of an element specified in text. In this example, the prompt is given by \code{The painting of a person wearing a beautiful <color> hat, <Image>}, with the colour of the hat specified below each image.}\label{fig:multimodal-inpainting}
\end{figure}

In pure text conditioned diffusion models, style modification can be achieved by starting from a corrupted version of an image and using a natural language instruction as a condition to steer the generation. The level of image corruption has to be chosen carefully, since there is a tradeoff between being close to the original image and enforcing the desired style. On the one hand, if only a little amount of noise is added to the image, the denoising process is very short and the conditional guidance has not enough steps to have a major effect. On the other hand, degrading the image too much one loses all information about it and the result is a random image in the desired style. In our experience, finding a sweetspot for the noise strength can be tricky for some combination of images and styles and the outputs are inconsistent. \\
As a byproduct of image synthesis with multimodal context, we present a new paradigm for style change using a multimodal prompt. We simply concatenate the style instruction with the image to form a single multimodal conditional prompt and then generate from pure noise. This way we can use the full diffusion length to steer generation while maintaining information about the original image consistently. A direct comparison of our new method to the old text-to-image approach with Stable Diffusion is depicted in Fig.~\ref{fig:multimodal-style}. In the same way we can also directly modify single image elements as illustrated in Fig.~\ref{fig:multimodal-inpainting}.
\subsection{Image composition}
\begin{figure}[t]
    \centering
    \includegraphics[scale=0.8]{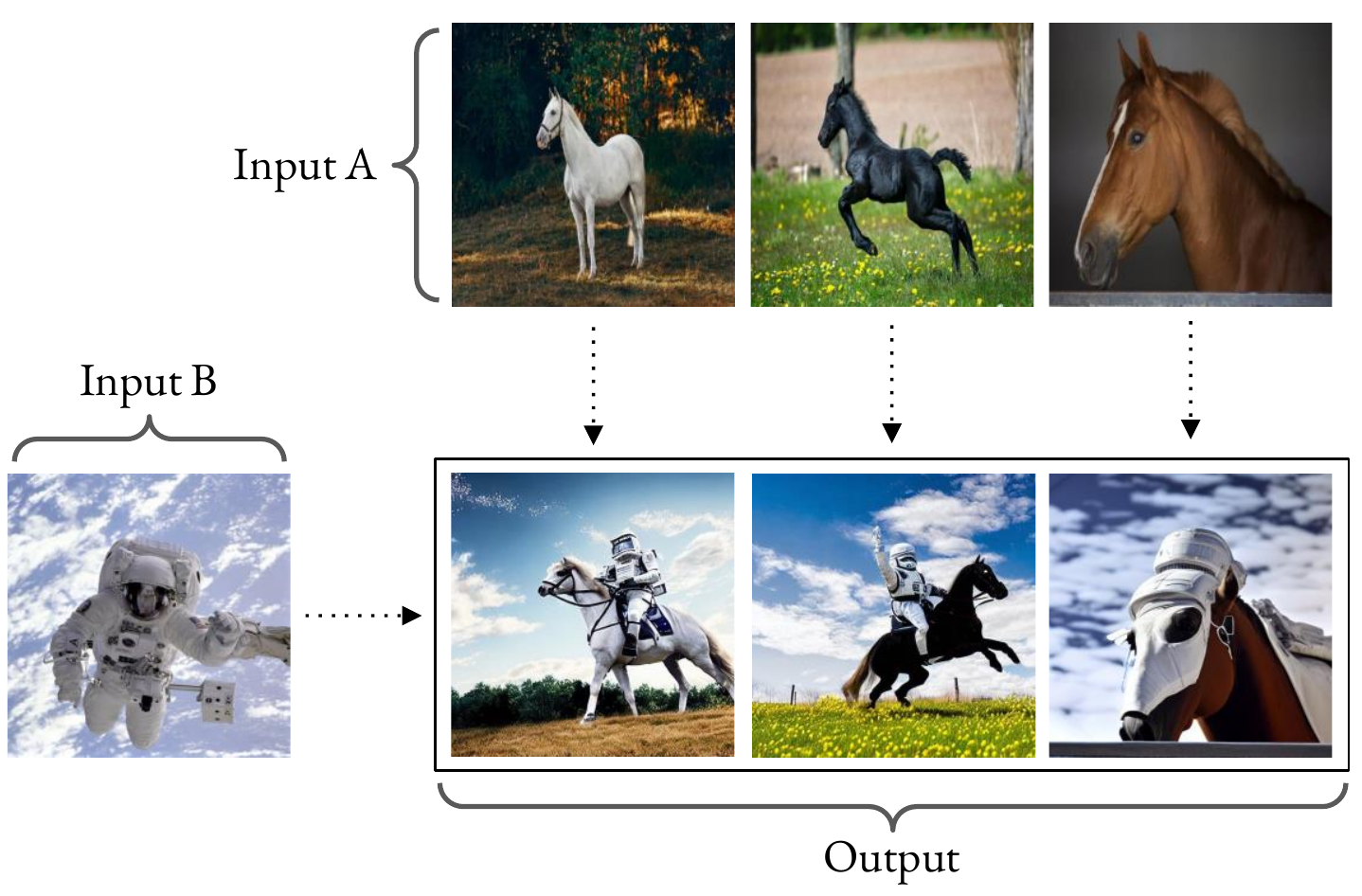}
\vspace{-5mm}
\caption{Combination of images for composition. M-VADER can meaningfully combine elements e.g. by having the astronaut ride the horse (left and center outputs), or, when the angle of the picture doesn't allow it, by turning the space suit into a space harness (right output).}\label{fig:image_composition}
\end{figure}

While the visual part of a CLIP encoder may directly be used to get variations of a base image~\cite{single_image_guided_stable_diffusion}, M-VADER can also out-of-the-box compose images with a prompt of the form
\begin{align*}
    \code{prompt: [<Image$_1$>, <Image$_2$>, \ldots].}
\end{align*}
In Fig.~\ref{fig:image_composition} we show an example of composition. The output image does not just reproduce the main elements of the prompt independently of one another, but also combines them in a complex and meaningful way. Note that this is conceptually very different from image interpolation in diffusion models, which works by adding a certain noise level to a set of images, then forming a convex combination of the noised latents and denoising again, resulting in a pixel space combination of the images rather than a combination in ``semantic space".

\section{Limitations and societal impact} \label{sec:limitations}

\textbf{Data limitations.}
Available large datasets of captioned images such as Laion 5B~\cite{laion} significantly shrink once filtered for unsafe content, minimal size and watermarks. In particular, the final set of images used for training stable diffusion and the model presented in this work consists of a modest amount of roughly 40M, which may be an obstacle to its generalization capabilities. We also found that in spite of the fact that stable diffusion has been trained on the unfiltered Laion 5B only for a small fraction of its full training, there's a significant probability of generating a watermarked image, depending on the prompt. See Appendix~\ref{sec:watermarks} for examples.
Aside for better or stricter filters, an interesting approach in this direction is to decrease the odds of generating a watermark in inference by the use of negative prompts.
Finally, our backbone model is trained on single pairs of images and text descriptions. Prior work has emphasised the importance of training AR models on multimodal sequences, i.e. on data sequences containing arbitrarily interleaved text, images and videos~\cite{cm3, flamingo}. We expect
our multimodal decoder, and the image diffusion capabilities as a consequence, to greatly benefit from training on such data, and leave to future work to estimate its impact.
\\

\textbf{Model limitations.} As this work is an extension of Stable Diffusion, we of course inherit limitations related to its architecture, e.g. concerning the use of latent diffusion for high resolution. We therefore refer to the original to~\cite{high_res_diffusion} for a more detailed discussion. Related to the use of a multimodal decoder, we observe that the content of generated images sometime depends on the order in which the images are concatenated.

\textbf{Ethical concerns.}
We identify misuse of the model and data as as the primary elements to consider for this work.
The field of generative modelling, specially of image distributions, is haunted by the threat of misuse, and this work is no exception. 
While it is clear how developments in the field are initiating a new era for creative content creation, research, and business applications, it is likewise apparent that countering a model's misuse gets harder and harder.
This becomes specially true with the continuous democratization of the field in terms of code/checkpoint availability and hardware resources needed to finetune a model or run inference with it.
We refer the interested reader to in depth studies on "deep fakes"~\cite{deep_fakes, NGUYEN2022103525}.

The final aspect to consider is the interaction between models and the data they are trained on. While we may cautiously consider training algorithms and to be unbiased, up to the fact that they were developed by humans, the same can't be said about data. \textit{"Biases in, biases out"}, i.e. it is well known~\cite{ethics2} that generative models can and do reproduce biases inherited from training data. Furthermore, the degree of memorization of a model make them prone to leak private data seen during training, limiting their application in data sensitive applications, e.g. in synthetic data generation for healthcare applications. We refer to~\cite{ethics1} for a more detailed breakdown of ethical concerns and additional sources on the topic.

\textbf{Quantitative evaluations.}
A standard and natural practice to evaluate text-to-image models would be to compute an FID score~\cite{FID} on the validation set of datasets like MS-COCO~\cite{coco}, conditioning on the provided caption. Unfortunately it is not clear how to quantitatively evaluate more complex tasks which M-VADER seems to perform well at. For example, for the task of style modification it would be preferable to evaluate on a dataset which provides triplets \code{(style instruction, input image, output image ground truths)}. Alternatively, style modification might be evaluated by using MS-COCO captions to retrieve images from an external pool based on a similarity score. However, doing so introduces an extra layer of complexity, as the alignment of the output with respect to the prompt would depend on the model used for search. Image composition presents a similar challenge, with no out-of-the box suitable datasets available (to our knowledge). We believe that evaluating complex image synthesis tasks is an important research topic in of itself with the potential to advance image generation models even further. We plan to address these challenges and properly evaluate M-VADER in future work.

\section{Conclusion}
In this work we have introduced M-VADER, a diffusion model capable of synthesizing images based on multimodal embeddings. We achieved this by assembling S-MAGMA, a 13B decoder-only transformer model combining components from a captioning model and biases finetuned for semantic search. We have displayed the capabilities of M-VADER using a variety of prompts, from generic sequences of visual and text elements, to subsets of it, such as image combinations and image variations. We see great potential for future developments in research on multimodal prompting, bringing us closer to image generation models that can better capture the intent of users for enhanced creative expression.

\section{Acknowledgements}
We are grateful to Stability AI for open sourcing the Stable Diffusion checkpoints and for the numerous details they provide to replicate the training set used in their work, as well as the hyperparameters used and details about the training procedure.

\bibliographystyle{ieeetr}
\bibliography{literature}

\newpage

\appendix

\newpage

\section{Image variations}\label{app:B}
\begin{figure}[h!]
    \centering
    \includegraphics[scale=0.45]{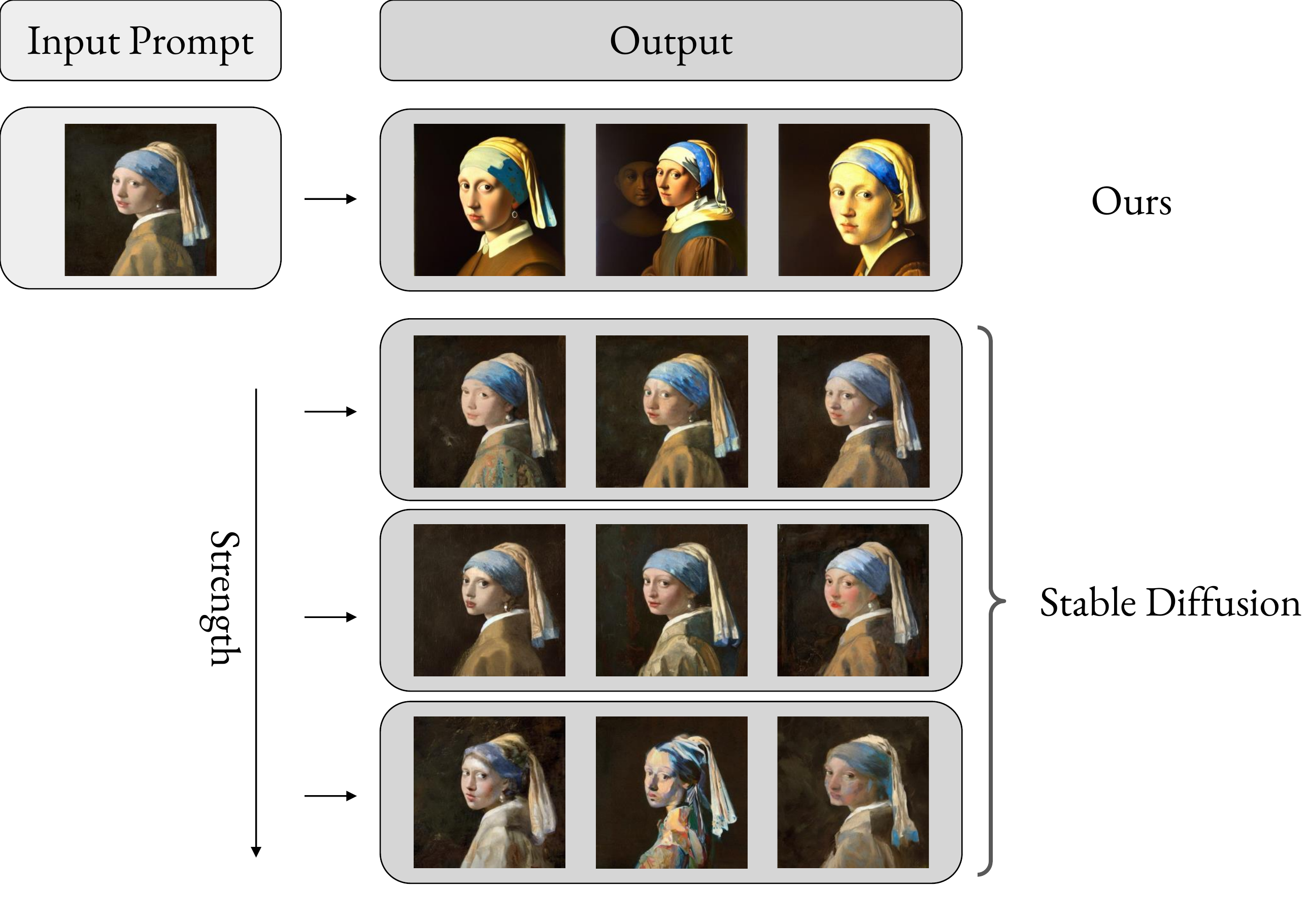}
\vspace{-5mm}
\caption{Three variations of a single image in two flavors. S-MAGMA embeds the input into a representation $H(I)$ which is then used to condition a denoising process from Gaussian noise via cross-attention. With a latent text-to-image model like Stable Diffusion, we start from a latent code of the image $z = Enc(I)$ and corrupt it with noise up to a time step $\tilde{T}$. $\tilde{T}$ needs to be tuned for each image, in such a way that the denoising of $z_{\tilde{T}}$ is not too close and not too far from $z$.}\label{fig:image_variations}
\end{figure}

M-VADER can out-of-the-box generate variations of an image by conditioning the diffusion model with an image embedding $H(I)$. While conceptually different, a somewhat similar result can be obtained by denoising a partially corrupt image with a vanilla diffusion model, with the corruption strength regulating the deviation from to original image. We provide in Fig.~\ref{fig:image_variations} a qualitative comparison of the two approaches.

\section{Watermarked outputs}\label{sec:watermarks}
\begin{figure}[h!]
    \centering
    \begin{tabular}{ccc}
    \centering
    \includegraphics[scale=0.2]{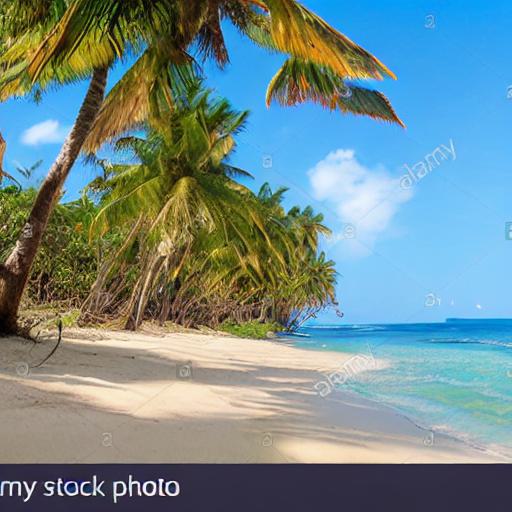}& 
    \includegraphics[scale=0.2]{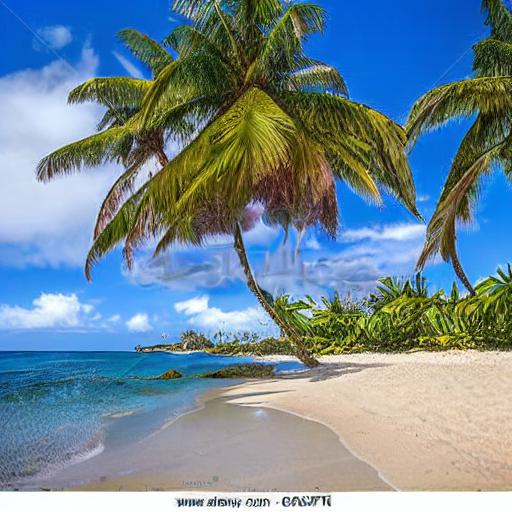}& 
    \includegraphics[scale=0.2]{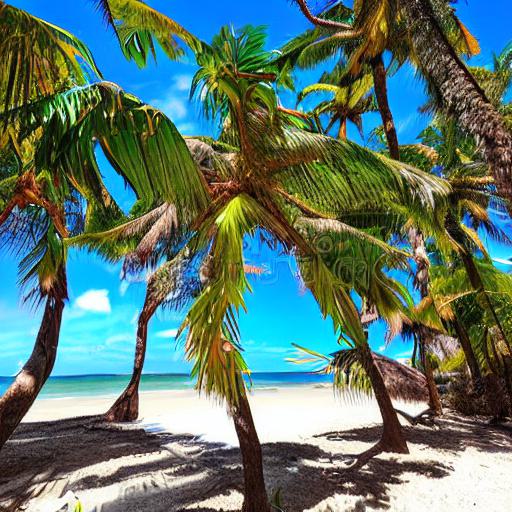}\\
    \includegraphics[scale=0.2]{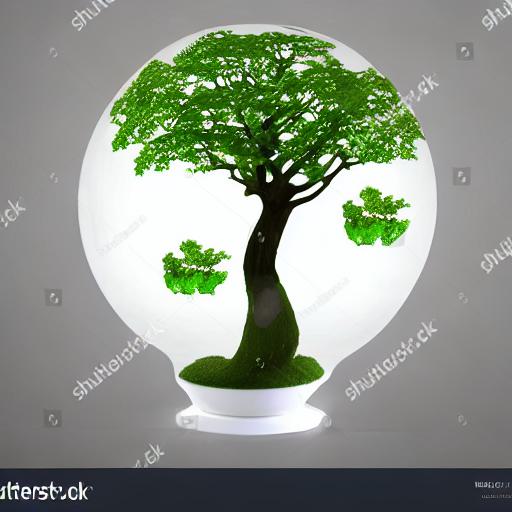}& 
    \includegraphics[scale=0.2]{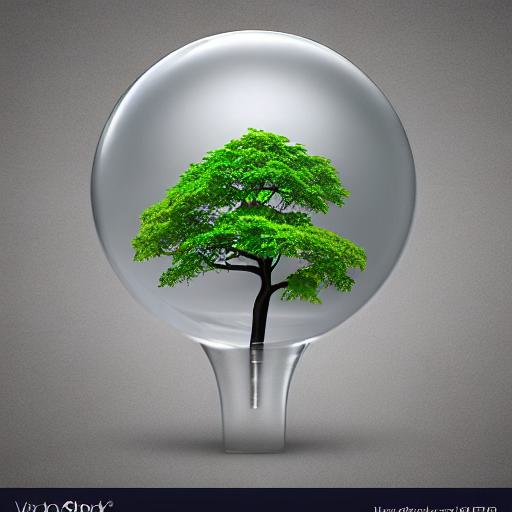}& 
    \includegraphics[scale=0.2]{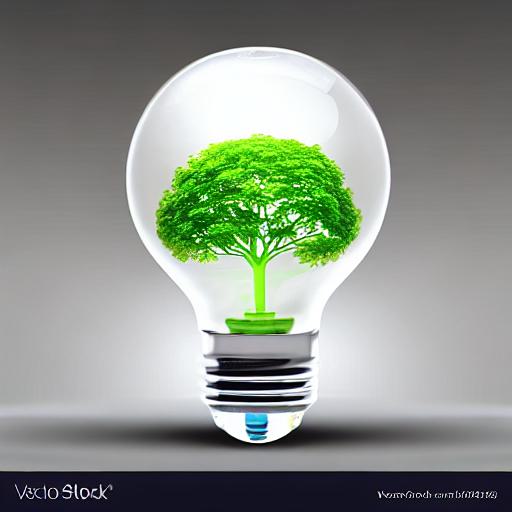}\\
    \end{tabular}
\caption{Examples of watermarked outputs. In the first row we condition the model over text with \code{prompt: coconut trees near the beach}. In the second row, we instead produce variations of an image (prompt can be visualized in Fig.~\ref{fig:appendix_variations}).}\label{fig:failure_mode_watermark}
\end{figure}

Following the original work of Stable Diffusion, we filtered out images with an estimated watermark probability $> 0.5$. However, we observe that some prompts have a significant chance of producing a watermarked output as shown in Fig.~\ref{fig:failure_mode_watermark}.
As per the authors description, the checkpoint used as starting point for this work has been pre-trained for $237$k steps on the much larger and much less filtered LAION 5B. This makes it hard to disentangle the source of watermarks training examples. While training on higher quality data is an attractive option, this will certainly come at the cost of slimming the training set even more, with unclear consequences on the generalization capabilities of the model. An alternative option is to accept instead the trade and use larger, less filtered data, provided that negative prompting is used in inference.

\section{Input order}
\begin{figure}[h!]
    \centering
    \includegraphics[scale=0.7]{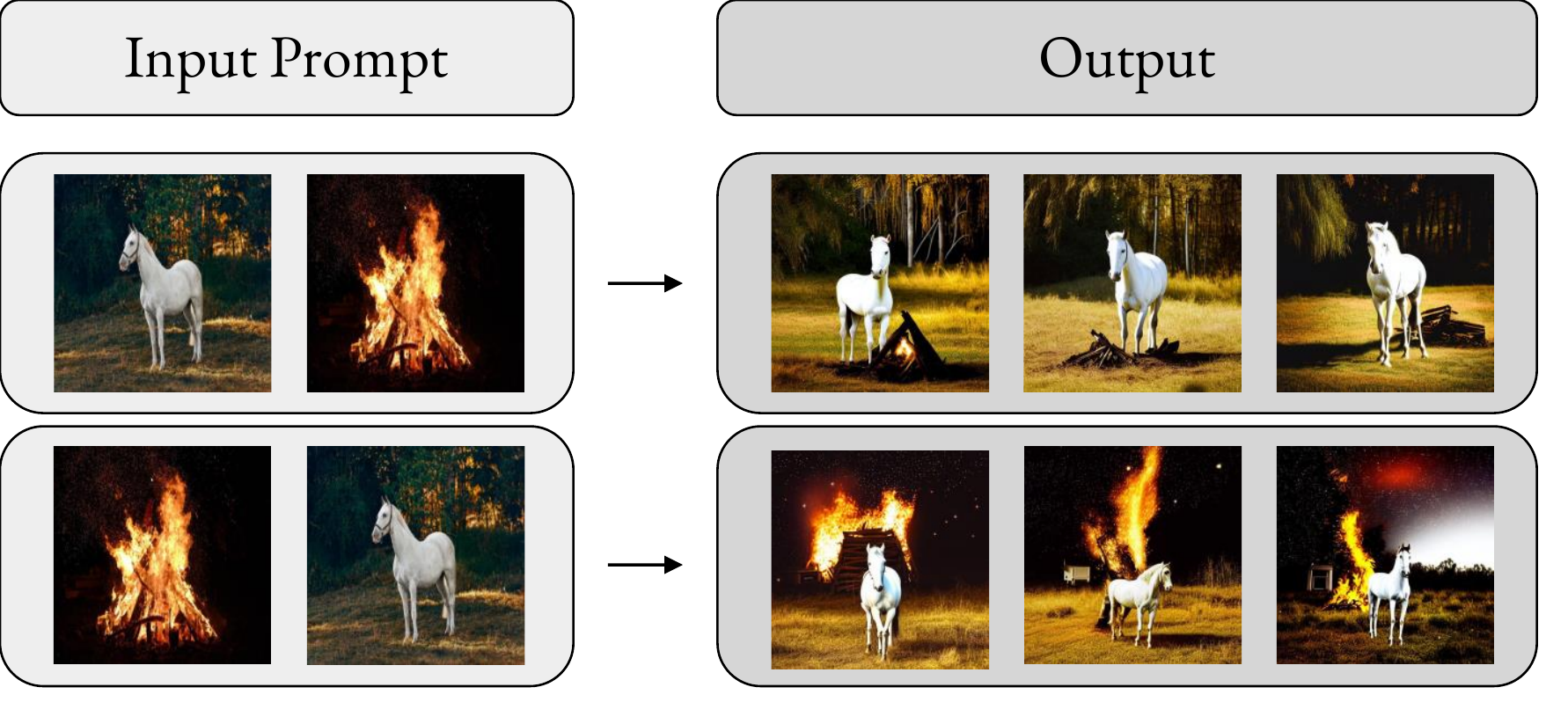}
\vspace{-5mm}
\caption{Image composition with the same pair of input figures, prompted in different orders}\label{fig:input_order}
\end{figure}

As the embeddings are generated autoregressively, the order of the elements in the prompt is more important with respect to models with bi-directional attention. This is particularly important for image composition, where no natural language prompt is used to specify the relative relationship between elements in the inputs. Since given a sequence of tokens information only flows causally in the forward direction, we expect the first element in the prompt to provide a stronger conditioning signal. We qualitatively verify this in Fig.~\ref{fig:input_order}: the background as well as the relative importance of the elements in the output is sensitive to the images order. 
\newpage

\section{Experiments details}
In this section we summarize the combinations of parameters finetuned during training as well as the relevant hyperparameters used in the experiments described in Sec.~\ref{subsec:training}.
\begin{table}[h!]
\centering
\scalebox{0.8}{
\begin{tabular}{cccc}
 \hline
 & UNET & Bias & MAGMA Adapters\\ 
 \hline
 \textbf{V1} & Finetuned & Sem. Search, finetuned & Not used\\ 
 \textbf{V2} & Finetuned & Sem. Search, frozen & Yes, frozen \\ 
 \hline
\end{tabular}}
\caption{Trained architecture variations}\label{tab:training_variants}
\end{table}
\begin{table}[h!]
\centering
\scalebox{0.8}{
\begin{tabular}{cccccc}
 \hline
 & training iters & lr cross attn & lr unet & batch size & embedding dropout rate \\ 
 \hline
    \textbf{V1} & 54k & 0.0005 & 5e-5 & 2048 & 0.05 \\ 
    \textbf{V2} & 60k & 0.0001 & 5e-5 & 2048 & 0.1 \\
 \hline
\end{tabular}}
\caption{Training details DDPM}\label{tab:training_diffuser}
\end{table}

\newpage

\section{Selection of generated images}
\begin{figure}[h!]
    \centering
    \includegraphics[scale=0.75]{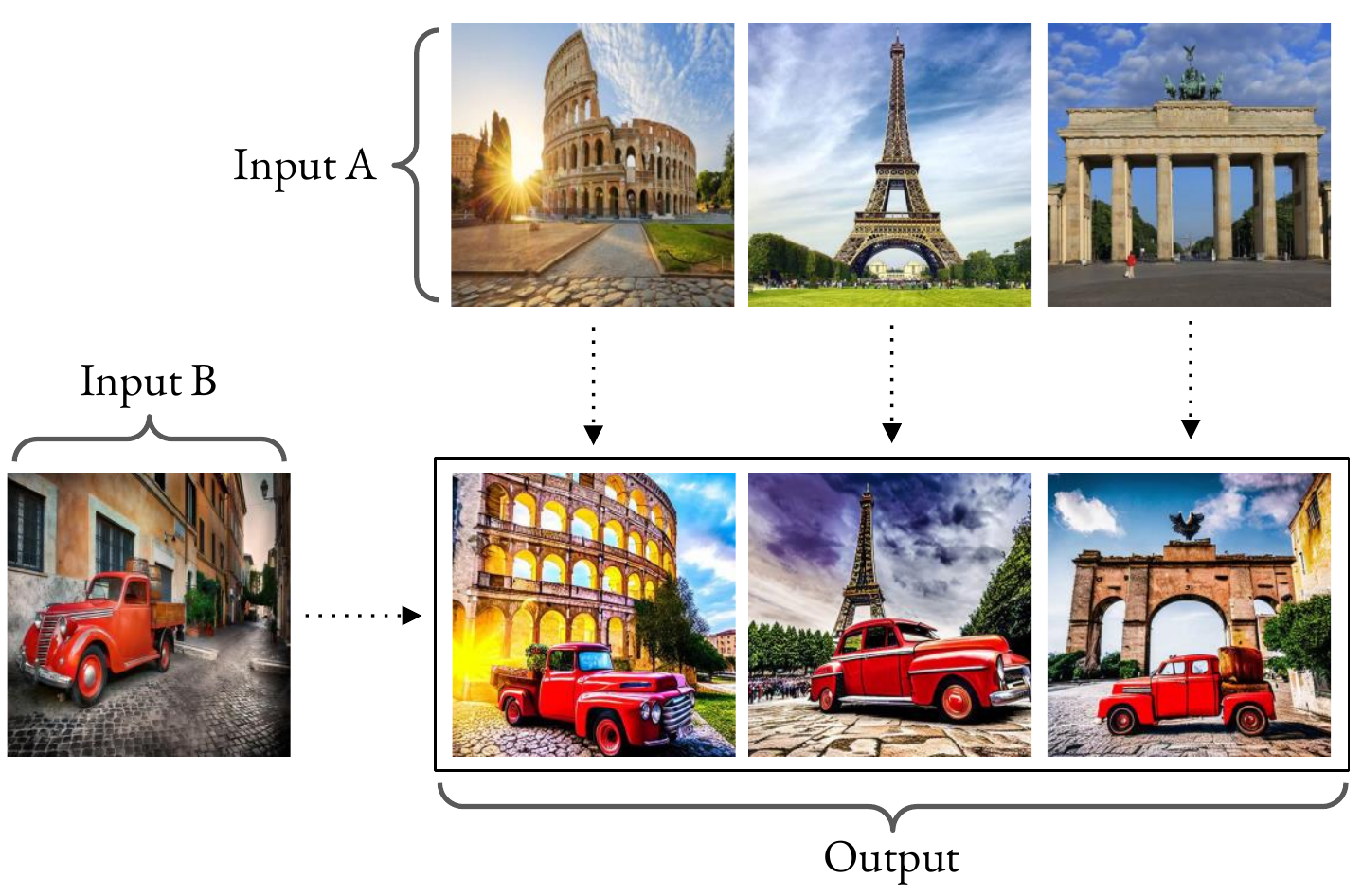}
    \vspace{-5mm}
\caption{Selection of compositions of image pairs.}\label{fig:appendix_composition}
\end{figure}

\begin{figure}[h!]
    \centering
    \includegraphics[scale=0.9]{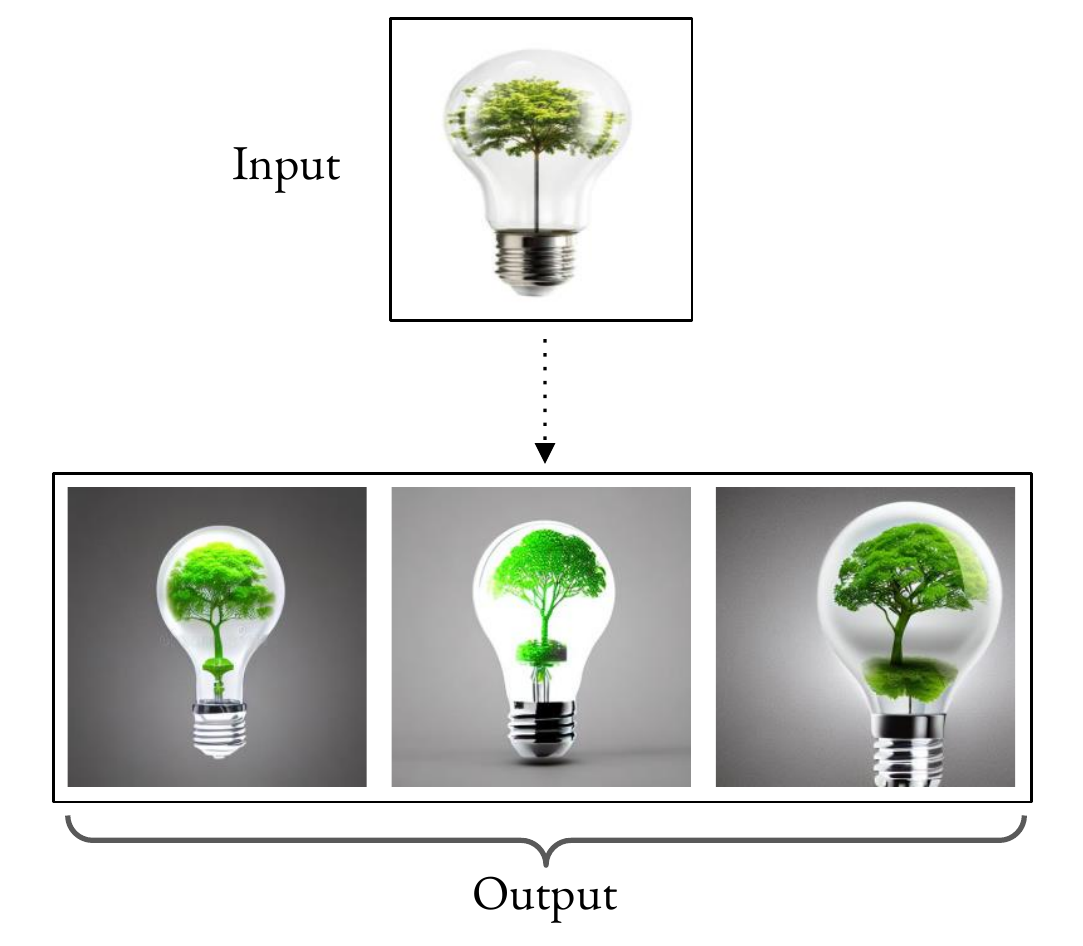}
    \vspace{-5mm}
\caption{Selection of variations of a base image.}\label{fig:appendix_variations}
\end{figure}

\begin{figure}[h!]
    \centering
    \includegraphics[scale=1.0]{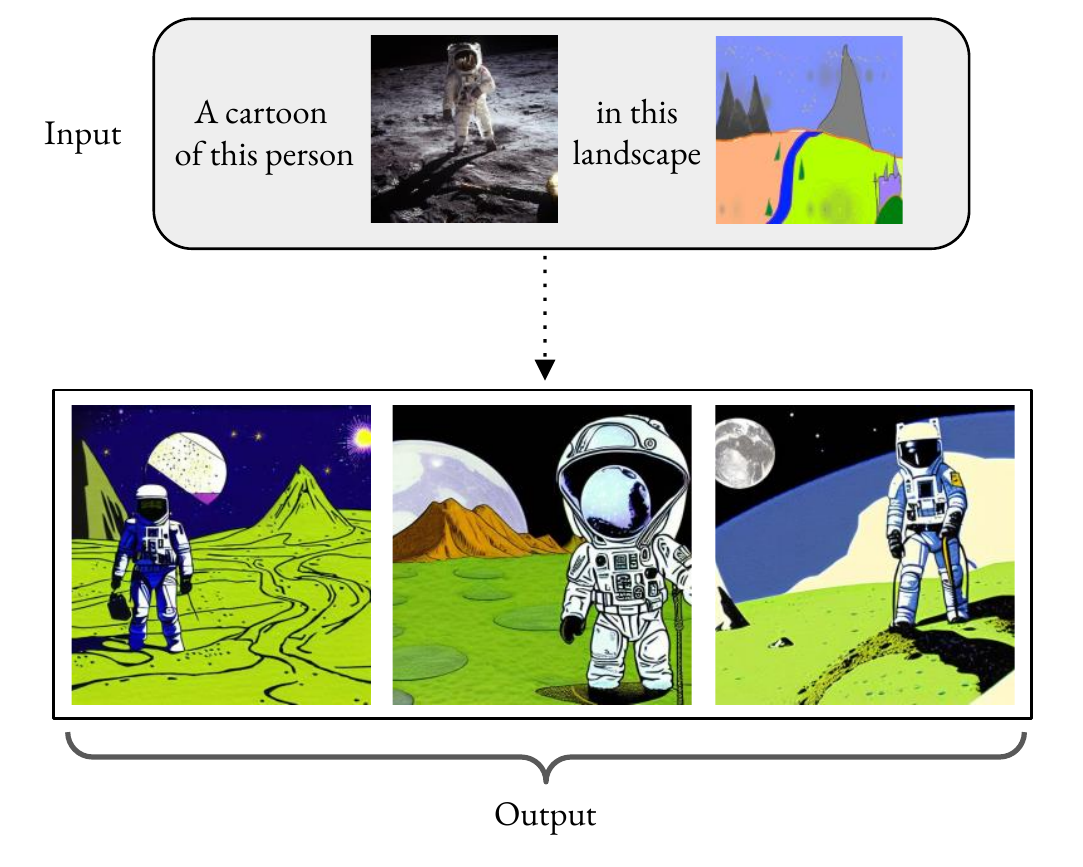}
\vspace{-5mm}
\caption{Selection of images synthesized with a multimodal prompt}\label{fig:appendix_multimodal_1}
\end{figure}

\begin{figure}[h!]
    \centering
    \includegraphics[scale=1.0]{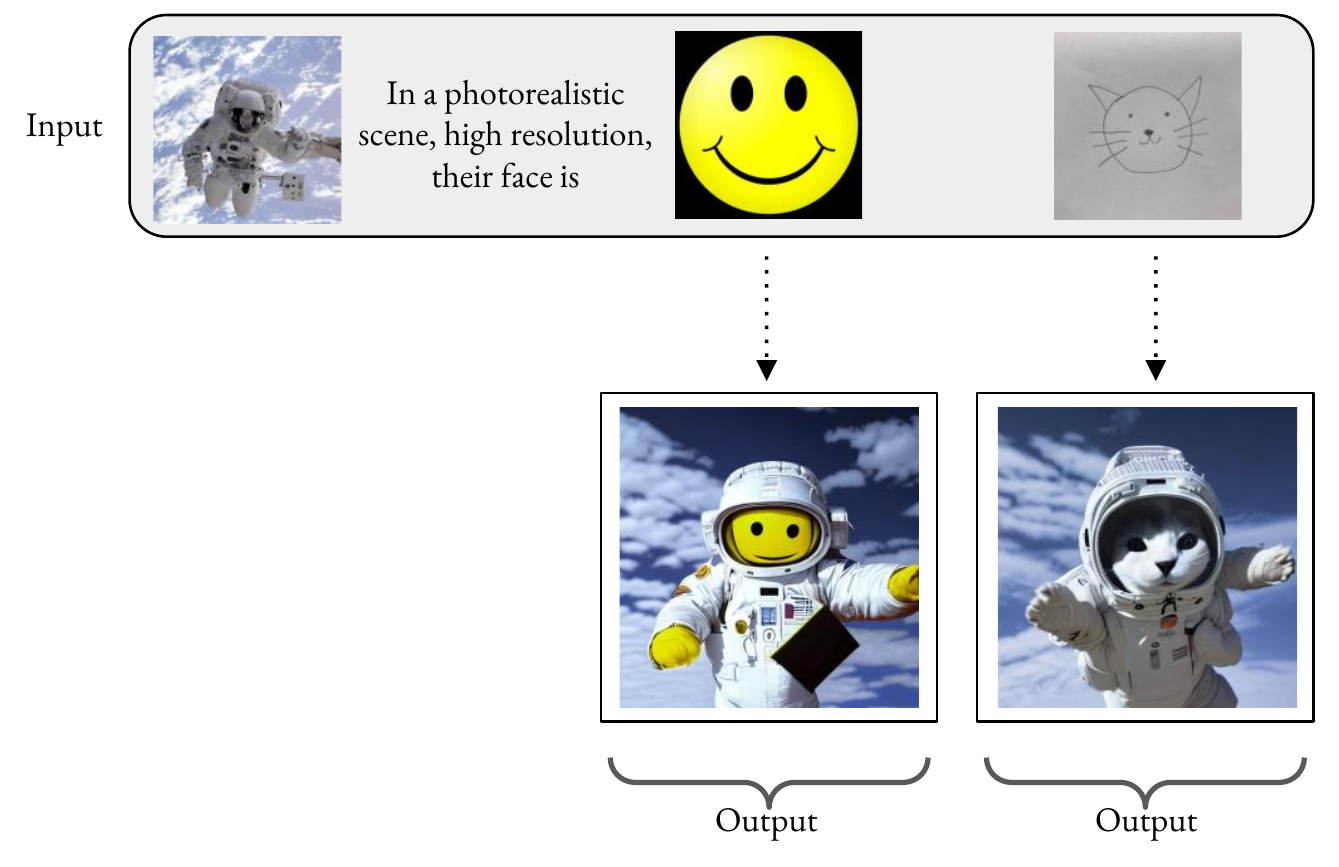}
\vspace{-5mm}
\caption{Selection of images synthesized with a multimodal prompt}\label{fig:appendix_multimodal_2}
\end{figure}

\begin{figure}[h!]
    \centering
    \includegraphics[scale=1.0]{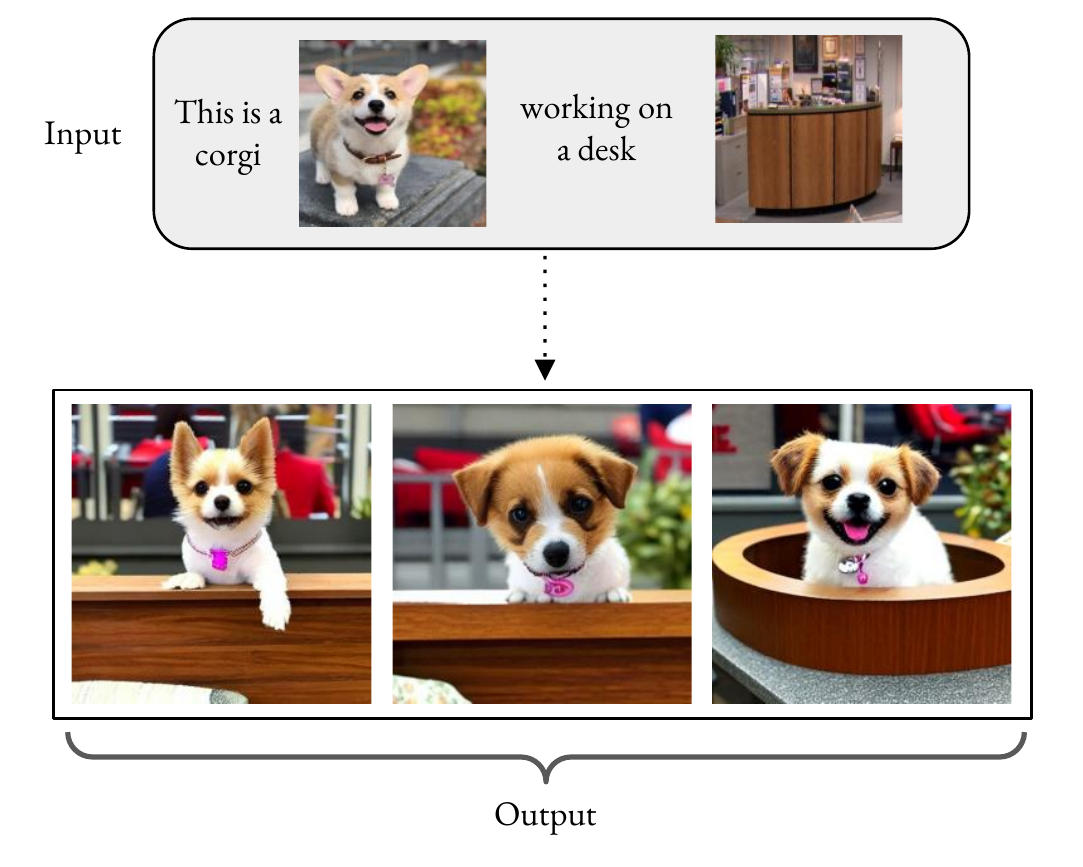}
\vspace{-5mm}
\caption{Selection of images synthesized with a multimodal prompt}\label{fig:appendix_multimodal_3}
\end{figure}

We present in Fig.~\ref{fig:appendix_composition}-Fig.~\ref{fig:appendix_multimodal_3} a selection of examples generated with M-VADER, spanning all the main use cases: multimodal prompts, images combination, image variations and style modification. 
\newpage
\newpage
\end{document}